\documentclass{article}

\usepackage{arxiv}
\usepackage[utf8]{inputenc}
\usepackage[T1]{fontenc}
\usepackage{hyperref}
\usepackage{url}
\usepackage{tabularx}
\usepackage{booktabs}
\usepackage{setspace}
\usepackage{amsfonts}
\usepackage{nicefrac}
\usepackage{microtype}
\usepackage{lipsum}
\usepackage{graphicx}
\usepackage{amsmath}
\usepackage{braket}
\usepackage{algorithm}
\usepackage{algorithmic}
\usepackage{orcidlink}
\graphicspath{ {./images/} }

\title{QMViT: A mushroom is worth 16x16 words}

\author{
 Siddhant Dutta \orcidlink{0009-0000-5120-7114} \\
  Department of Computer Engineering \\
  Dwarakdas J. Sanghvi College of Engineering \\
  Mumbai, India 400056 \\
\texttt{forsomethingnewsid@gmail.com} \\
   \And
 Hemant Singh \orcidlink{0009-0009-0437-8688}\\
  Department of Computer Engineering \\
  Dwarakdas J. Sanghvi College of Engineering \\
  Mumbai, India 400056 \\
\texttt{hphemant373@gmail.com} \\
  \And
 Kalpita Shankhdhar \\
  Department of Computer Engineering \\
  Dwarakdas J. Sanghvi College of Engineering \\
  Mumbai, India 400056 \\
  \texttt{kalpitashankhdhar@gmail.com}
   \And
 Sridhar Iyer \\
  Department of Computer Engineering \\
  Dwarakdas J. Sanghvi College of Engineering \\
  Mumbai, India 400056 \\
  \texttt{sridhar.iyer@djsce.ac.in}
}

\begin{document}
\maketitle
\begin{abstract}
	
	Consuming poisonous mushrooms can have severe health consequences, even resulting in fatality and accurately distinguishing edible from toxic mushroom varieties remains a significant challenge in ensuring food safety. So, it's crucial to distinguish between edible and poisonous mushrooms within the existing species. This is essential due to the significant demand for mushrooms in people's daily meals and their potential contributions to medical science. This work presents a novel Quantum Vision Transformer architecture that leverages quantum computing to enhance mushroom classification performance. By implementing specialized quantum self-attention mechanisms using Variational Quantum Circuits, the proposed architecture achieved 92.33\% and 99.24 \% accuracy based on their category and their edibility respectively. This demonstrates the success of the proposed architecture in reducing false negatives for toxic mushrooms, thus ensuring food safety. Our research highlights the potential of QMViT for improving mushroom classification as a whole.
	
\end{abstract}

\section{Introduction}

Mushrooms, a popular food item worldwide, have numerous types with varying toxicity levels, making it difficult for people to identify the types and determine which ones are safe for consumption, especially because some of them can be highly toxic and cause serious health issues, including death \cite{wang2020mushroom}. 
Quantum computing is a new technology that encodes information using quantum mechanical functions such as superposition and quantum entanglement, which are absent in traditional computers. This technology provides quicker computations, the ability to solve a greater range of problems and improved performance for machine learning applications \cite{soares2023quantum, quant}. When used properly, these traits can enable novel solutions to challenging problems \cite{golestan2023quantum}. However, contemporary quantum computers have a finite number of qubits, which represent the superposition of two quantum states, and are prone to noise, an inherent quantum mistake \cite{yan2014quantum}. Preskill \cite{preskill2018quantum} coined the term "Noisy Intermediate Scale Quantum (NISQ) Era" to describe the current stage of quantum computing development.

Quantum Machine Learning (QML), a newly emerging field, combines classical machine learning techniques with the unique capabilities of quantum computing. This integration presents promising avenues for addressing complex challenges across diverse domains \cite{peruzzo2014variational, du2020expressive}. In the context of mushroom classification, QML offers an approach harnessing the distinctive features of quantum systems, to tackle the difficulties posed by the immense global diversity of mushrooms \cite{cerezo2022variational, maheshwari2021variational}. This quantum-based methodology holds significant potential to enhance the accuracy and efficiency of identifying mushroom types and predicting their toxicity levels leading to new approaches in ensuring safety in public health.

\subsection{Poisonous Mushrooms on a World Scale}
Mushrooms, a global food source, have a diverse range, including many highly toxic species. The prevalence of poisonous mushroom species across different continents, highlighting the estimated number of poisonous species according to the available data, is as follows:

\begin{itemize}
	\item \textbf{North America:} According to estimates, North America has over 500 species of poisonous mushrooms, including the infamous Amanita phalloides (death cap), the destroying angel, the deadly galerina, and the deadly webcap, which can cause severe illness or death on consumption.
	\item \textbf{Europe:} Estimates suggest Europe has around 600 documented poisonous mushroom species. Foraging for wild mushrooms is popular in Europe, leading to a higher incidence of mushroom poisoning. The death cap, the fool’s mushroom, the panther cap, and the sweating mushroom are the major toxins, causing a range of symptoms, from gastrointestinal distress to liver and kidney failure.
	\item \textbf{Asia:} is estimated to have over 1000 species of poisonous mushrooms. The death cap, the fly agaric that contains psychoactive compounds, and the deadly shiitake which is a look-alike to its edible counterpart are present. 
	\item \textbf{Australia:} Estimates suggest that there are around 300 species of poisonous mushrooms in Australia, with the death cap, the yellow stainer, the funeral bell and the ivory funnel-cap being prominent threats. Some Australian fungi contain amatoxins, the same deadly toxins found in the death cap, highlighting the importance of proper identification.
	\item \textbf{South America:}  It is estimated that around 400 species of poisonous mushrooms are present in South America. The death cap, the destroying angel, muscarine, and the false chanterelle are dangerous and can be easily mistaken for its edible counterpart.
	\item \textbf{Middle East:} Estimates suggest that there are over 150 potential threats in this region, including the death cap, the destroying angel, the yellow stainer, and the fly agaric being present in abundance.
\end{itemize}

Combining classical machine learning techniques with the capabilities of quantum computing, through the emerging field of Quantum Machine Learning (QML), holds promise for addressing these challenges of detection. QML-based approaches leverage the distinctive properties of quantum systems to enhance the accuracy and efficiency of mushroom classification which in turn provides a critical tool for safeguarding public health in high-risk areas.

\subsection{Types of Toxins and Symptoms in Mushroom Poisoning}
Mushroom poisoning is a significant public health concern, encompassing a wide range of adverse reactions caused by the ingestion of toxins present in various fungal species. These toxins can be classified into several categories, each with distinct chemical structures and physiological effects \cite{mishra2019mushroom}. The major types of toxins found in poisonous mushrooms and their associated symptoms are:

\subsubsection{Amatoxins}
\begin{itemize}
	\item A group of bicyclic peptides, these are the most dangerous toxins found in mushrooms.The principal amatoxins include $\alpha$-amanitin and $\beta$-amanitin \cite{bever2020rapid}. They are primarily present in mushrooms belonging to the genus Amanita, most notably the infamous death cap (Amanita phalloides). The symptoms of Amatoxin poisoning follow a biphasic course where the first phase (6-24 hours) can be misleading, with symptoms like gastrointestinal distress (nausea, vomiting, diarrhoea) that may subside within a day and then the second phase (24-72 hours) which has a deceptive period of apparent improvement that can be followed by severe liver and kidney damage, leading to coma and death if left untreated.
\end{itemize}

\subsubsection{Gastrointestinal Irritants}
\begin{itemize}
	\item The various complex carbohydrates, proteins, and amino acids present in certain mushrooms can act as gastrointestinal irritants. These toxins are not well-characterized at the molecular level and can be found in species from genera like Agaricus, Lepiota, Boletus (few species), and Paxillus \cite{mishra2019mushroom}. The symptoms typically appear within a few hours of ingestion and are usually mild to moderate. They may include nausea, vomiting, abdominal cramps, and diarrhoea. In severe cases, dehydration and electrolyte imbalance can necessitate medical attention.
\end{itemize}

\subsubsection{Muscarine}
\begin{itemize}
	\item It is a low-molecular-weight alkaloid that disrupts the parasympathetic nervous system. It is found in a limited number of mushroom species, including Omphalotus illudens, Clitocybe dealbata, and some Inocybe species.\cite{avila2020development}. The symptoms of Muscarinic poisoning usually begin within 15-30 minutes of ingestion. The characteristic symptoms include excessive salivation, sweating, tears, diarrhoea, and visual disturbances. In severe cases, respiratory failure can occur, that might be fatal.
\end{itemize}

\subsubsection{Isoxazole Derivatives (Muscimol and Ibotenic Acid)}
\begin{itemize}
	\item These toxins are glutamate receptor agonists, causing neuroexcitation and central nervous system effects. Isoxazole derivatives are present in certain Amanita species, most notably the fly agaric (Amanita muscaria).\cite{avila2020development}. The symptoms typically appear within 30 minutes to 2 hours of ingestion and can include confusion, hallucinations, drowsiness, and convulsions. Fatalities are uncommon with isoxazole derivative poisoning.
\end{itemize}

Several other types of toxins can be found in poisonous mushrooms, each with varying effects. These include gyromitrin, which causes effects similar to vitamin B6 deficiency; orellanine, which damages the kidneys; psilocybin and psilocin, which produce hallucinations; and coprine, which causes unpleasant reactions when consumed with alcohol. 

\subsubsection{Current Standard Ways to Avoid Mushroom Poisoning}
Mushroom poisoning poses a significant health risk globally. To mitigate this risk, several standard practices have been established to help individuals identify and avoid toxic mushrooms. These practices draw upon chemical knowledge and practical experience to ensure safe foraging and consumption of wild fungi \cite{mishra2019mushroom}.
\begin{enumerate}
	\item \textbf{Species Identification:} Accurate identification of mushroom species is paramount to avoiding poisoning. Utilize comprehensive field guides and consult with mycologists to correctly identify mushrooms. Pay close attention to distinguishing features such as cap shape, gill attachment, spore color, and the presence of a volva or ring. \cite{smith1875mushrooms}
	      	      	      	          
	\item \textbf{Chemical Analysis:} Chemical tests can aid in the identification of toxic mushrooms. For example, the presence of alpha-amanitin, a lethal amatoxin found in certain Amanita species such as \textit{Amanita phalloides}, can be detected using enzyme-linked immunosorbent assay (ELISA) kits. Employing these tests can provide an additional layer of certainty in mushroom identification. \cite{mishra2019mushroom}
	      	      	      	          
	\item \textbf{Spore Print Examination:} Obtaining a spore print, which involves placing a mushroom cap on a piece of paper to capture its spores, can assist in species identification. Different mushroom species produce spores of varying colors, aiding in classification. For instance, \textit{Amanita muscaria} typically yields a white spore print, while \textit{Coprinopsis atramentaria} produces a black spore print. \cite{mishra2019mushroom}
	      	      	      	          
\end{enumerate}

\section{Related Works}
Academia and the public at large have shown a great deal of interest in the study of mushroom poisoning detection and prevention. Numerous investigations have examined diverse methodologies, spanning from conventional techniques to cutting-edge technologies, to tackle the obstacles linked to the toxicity of mushrooms. This section reviews important works on the topic of fungal poisoning prevention and detection, emphasizing both modern developments and historical contributions.

\subsection{Historical Perspective}
Early efforts in mushroom poisoning research primarily relied on empirical observations, taxonomic classification, and anecdotal evidence. The pioneering work of mycologists such as Elias Magnus Fries and David Arora laid the foundation for modern mycology and contributed to the identification of poisonous mushroom species. Historical studies often documented poisoning cases described symptoms and proposed rudimentary guidelines for mushroom foragers to avoid toxic species.

One of the seminal works in mushroom poisoning literature is "Mushrooms and Toadstools: How to Distinguish Easily the Differences Between Edible and Poisonous Fungi" by W. George Smith, published in 1875 \cite{smith1875mushrooms}. 
This comprehensive guide provided practical advice on identifying edible and poisonous mushrooms based on morphological characteristics, spore prints, and habitat preferences. Although lacking in scientific rigor compared to contemporary standards, Smith's work played a crucial role in promoting public awareness of mushroom toxicity.

\subsection{Chemical Analysis and Toxin Identification}
In addition to image-based approaches, researchers have explored chemical analysis techniques for detecting and quantifying mushroom toxins. High-performance liquid chromatography (HPLC), gas chromatography-mass spectrometry (GC-MS), and enzyme-linked immunosorbent assays (ELISA) are commonly employed methods for analyzing mushroom samples and identifying specific toxins \cite{bever2020rapid}. 

For example, Parant et al. (2006) \cite{parant2006phalloidin} developed an ELISA-based assay for detecting alpha-amanitin, a potent hepatotoxin found in deadly Amanita species.
The assay exhibited high sensitivity and specificity, enabling rapid and accurate detection of alpha-amanitin in mushroom extracts. Similarly, Avila et al. (2020) \cite{avila2020development} utilized 
HPLC to quantify various toxins, including muscarine and ibotenic acid, in mushroom samples, providing valuable insights into the chemical composition of toxic mushrooms.

\subsection{Machine Learning Approaches}
Recently, machine learning (ML) and artificial intelligence (AI) techniques have emerged as useful tools for the detection and classification of fungal toxins. Several studies have used ML algorithms such as support vector machines (SVMs), random forests \cite{wang2020mushroom}, and convolutional neural networks (CNNs) to analyze fungal images and predict toxicity levels \cite{zahan2021deep}. 
For example, Yagmur et al. (2023) \cite{demirel2023deep} developed a deep learning model that can identify poisonous mushrooms with high accuracy based on image data. The model uses a CNN architecture trained on a large dataset of labeled mushroom images to achieve better performance compared to existing image processing methods. Similarly, Doung et. al (2023) \cite{duong2023ensemble} proposed a gradient boosting framework to classify fungal species and predict toxicity using morphological features extracted from images. 
Many studies have used convolutional neural networks (CNNs) \cite{ketwongsa2022new}, which are part of deep learning (DL) models, to classify fungi based on image data \cite{preechasuk2019image}. For example, Ottom et al. \cite{ottom2019classification} used various machine learning techniques including neural networks (NN), support vector machines (SVM), decision trees, and nearest neighbors (kNN) to classify fungal images taken from public datasets. In this way, kNN achieved 94\% accuracy in extracting image features and sizes of different fungi. Wagner et al. \cite{wagner2021mushroom} collected a large dataset to predict mushroom consumption and evaluated various machine learning models such as Naive Bayes, logistic regression, linear discriminant analysis, and random forest (RF). The RF model showed excellent performance with 100\% accuracy using 5-fold cross-validation. Emine Cengil et al. \cite{cengil2021poisonous} prepared a dataset representing the eight most toxic fungal species and adjusted the YOLOv5 algorithm for their recognition. Their model achieved a perfect accuracy of 0.77. 
In recent years, DL models have become increasingly popular in fungal classification tasks due to their ability to automatically extract features from images \cite{lecun2015deep}. For example, Sajedi et al.  \cite{sajedi2019automated} achieved an accuracy of 80.7\% with the CNN-MLP model using a four-layer CNN base to identify mucoid taxa. Devika et al. \cite{devika2021identification} proposed a mushroom classification system based on deep convolutional neural network (DCNN), which outperforms machine learning interpreters. Wang et al. \cite{wang2021classification} introduced a bilinear CNN (B-CNN) and a sensitivity method for Amanita classification and achieved an accuracy of 95.2\% on the test set. 
Despite progress in DL-based classification of fungi, challenges remain, including the need for characterization and characterization of DL samples.

\section{Materials \& Methods}

We approach the task of toxic mushroom detection as a species classification problem. Our dataset includes information on mushroom characteristics and habitat. We gather data from medical journals and conduct data cleaning and preprocessing steps.

\subsection{Description \& Practical Use of the Dataset}

The Kaggle dataset \cite{almaMush} presents a comprehensive collection of over 50,000 meticulously curated, high-resolution photographs showcasing 100 distinct species of mushrooms indigenous to the Russian Federation. Sourced from iNaturalist, a renowned platform dedicated to documenting global biodiversity, with a particular emphasis on fungi and lichens, this dataset offers researchers and enthusiasts alike a valuable resource for exploring the rich diversity of mushroom species.

\begin{figure}[h]
	\centering
	\includegraphics[height=0.475\linewidth]{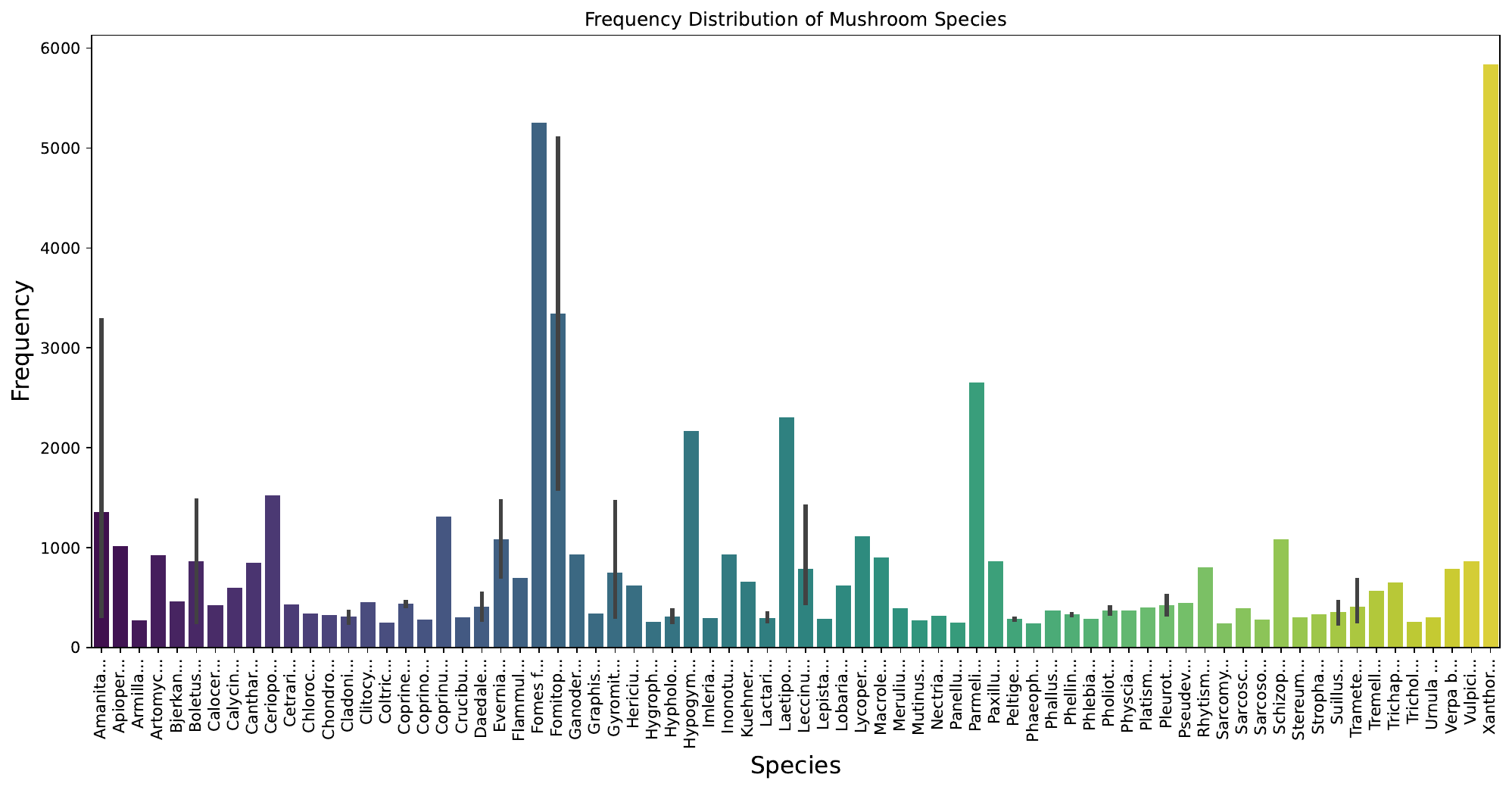}
	\caption{Distribution of labels within the multi-class dataset}
\end{figure}

Each species within the dataset is fully labelled with its corresponding scientific name, providing a robust taxonomy for the mushrooms represented. This taxonomic range covers a wide array of fungal classifications, from well-known genera like \textit{Amanita}, \textit{Boletus}, and \textit{Cantharellus} to lesser-known species.

The practical applications of this dataset are manifold, spanning various domains:

\begin{itemize}
	\item \textbf{Mycological Research:} Researchers can use this dataset to investigate the morphological characteristics, distribution patterns, and ecological roles of various mushroom species. Through image analysis and associated metadata, insights into fungal biodiversity, taxonomy, and evolutionary relationships can be extracted, improving our understanding of them.
	      	      	      	          
	\item \textbf{Species Identification:} The dataset is a useful resource for training and validating machine learning models and computer vision algorithms to identify and classify mushroom species. These models can simplify the process of species identification, especially in field situations where manual identification is impossible and time-consuming.
	      	      	      	          
	\item \textbf{Biodiversity Assessment:} Conservationists and environmental scientists can utilize the dataset to assess fungal biodiversity across diverse regions and habitats. Analysis of the abundance and distribution of mushroom species depicted in the images can aid in monitoring ecosystem health, identifying biodiversity hotspots, and tracking changes in fungal communities over time.
\end{itemize}

\subsection{Preprocessing of Dataset}

The preprocessing of the dataset involves several steps to prepare the images for input into the deep learning models.

Firstly, the initial step involves defining the transformations to be applied to the input images during training and validation. For training data, the following transformations are applied:
\begin{itemize}
	\item \textbf{Resize:} The images are resized to the specified input size.
	\item \textbf{RandomRotation:} Random rotation up to 20 degrees is applied to augment the training data.
	\item \textbf{RandomAdjustSharpness:} Random adjustment of sharpness is performed to further augment the data.
	\item \textbf{ToTensor:} The images are converted to tensors.
	\item \textbf{Normalize:} Image normalization is applied using the calculated mean and standard deviation.
	      	      	      
	      \begin{enumerate}
	      	\item \textbf{Mean and Standard Deviation Calculation:} First, the mean and standard deviation values are calculated for each channel of the input image dataset. This is typically done by computing the mean and standard deviation across all pixel values of each channel in the entire dataset.
	      	      	      	      	      	          
	      	\item \textbf{Per-Channel Normalization:} Once the mean ($\mu$) and standard deviation ($\sigma$) are calculated, the normalization is performed independently for each channel of the image tensor. For a given pixel value $x_{ijc}$ at location $(i, j)$ in channel $c$, the normalized pixel value $\hat{x}_{ijc}$ is calculated using the following equation:
	      	      	      	      	      	          
	      	      \begin{equation}
	      	      	\hat{x}_{ijc} = \frac{x_{ijc} - \mu_c}{\sigma_c}
	      	      \end{equation}    
	      	      Where:
	      	      \begin{itemize}
	      	      	\item $x_{ijc}$ is the pixel value of the input image at location $(i, j)$ in channel $c$.
	      	      	\item $\mu_c$ is the mean value of channel $c$ calculated across the entire dataset.
	      	      	\item $\sigma_c$ is the standard deviation value of channel $c$ calculated across the entire dataset.
	      	      \end{itemize}
	      	      	      	      	      	          
	      	\item \textbf{Normalization Result:} After normalization, each channel of the image tensor has a mean of approximately 0 and a standard deviation of approximately 1. This normalization process helps in reducing the effects of variations in lighting conditions and contrasts across different images, making the training process more stable and efficient.
	      \end{enumerate}
	      	      	       
\end{itemize}

The validation data undergoes resizing and normalization transformations, without any additional augmentation. Subsequently, these preprocessing steps convert the images into a 3-channel format and apply the specified transformations, resulting in the processed examples.

\subsection{Suggested Metrics}

The evaluation of classification algorithms can be done in various ways. Sometimes, metrics that measure the same thing have different names depending on the discipline in which they are discussed. However, irrespective of nomenclature, two critical considerations prevail. Firstly, sensitive species datasets often exhibit class imbalances, where one class dominates the dataset. Addressing this, metrics should be appropriately weighted based on class distribution, which is particularly crucial in binary classifications. Secondly, presenting an expansive array of outcome metrics beyond the conventional subset of accuracy and F1-score is essential for a comprehensive assessment.\cite{metricsmush}

In light of these considerations, the following metrics are suggested for evaluating classification algorithms applied to the mushroom species dataset:

\begin{itemize}
	\item \textbf{Accuracy}: The proportion of correctly classified instances out of the total instances. \cite{metricsmush}
	\item \textbf{Precision}: The proportion of true positive predictions out of all positive predictions made by the model. \cite{metricsmush}
	\item \textbf{Recall (Sensitivity)}: The proportion of true positive predictions out of all actual positive instances in the dataset. \cite{metricsmush}
	\item \textbf{F1-score}: The harmonic mean of precision and recall, providing a balanced measure of a model's performance. \cite{metricsmush}
	\item \textbf{Specificity}: The proportion of true negative predictions out of all actual negative instances in the dataset. \cite{metricsmush}
	\item \textbf{ROC-AUC (Receiver Operating Characteristic - Area Under the Curve)}: The area under the ROC curve, which plots the true positive rate against the false positive rate. It provides a measure of a model's ability to discriminate between positive and negative classes. \cite{metricsmush}
	\item \textbf{PR-AUC (Precision-Recall Area Under the Curve)}: The area under the precision-recall curve, provides insight into a model's performance across different thresholds. \cite{metricsmush}
	\item \textbf{Balanced Accuracy}: The average of sensitivity and specificity, accounting for class imbalances in the dataset. \cite{metricsmush}
	\item \textbf{Matthews Correlation Coefficient (MCC)}: A correlation coefficient between the observed and predicted binary classifications, suitable for imbalanced datasets. \cite{mccmush}
\end{itemize}

These metrics offer a comprehensive view of a model's performance, considering factors such as class imbalances, discrimination ability, and overall predictive accuracy.  To produce more robust and generalizable models, we also recommend implementing a cross-validation strategy.

\subsection{Machine Learning Models}

Machine learning models such as NuSVC and XGBClassifier have been widely used for classification tasks, including mushroom species identification. These models employ different approaches to delineate decision boundaries in the feature space, enabling them to classify instances into distinct classes.

Gao et al. \cite{gao2020classification} proposed a mixed FT-IR scans and machine learning techniques for detecting Bachu mushrooms and distinguishing them from other fungi. This emphasis on Bachu mushrooms is owing to their great nutritional and therapeutic worth, making them a sought-after product on the market. Employing machine learning algorithms such as Support Vector Machine (SVM), Backpropagation Neural Network (BPNN), and k-nearest neighbors (KNN) to process the spectral data. SVM, noted for its generalization performance, was employed for direct classification of mushroom powder following feature extraction. BPNN, chosen for its great nonlinear mapping capabilities was applied for multivariate classification. The classification accuracy of the models varied based on the amount of Partial Least Squares (PLS) components used. with instance, the BPNN algorithm achieved an accuracy of 99.07 percentage with caps data with 25 PLS components and 98.70 percentage for stalks data.

\subsubsection{NuSVC (Support Vector Classification)}

Support Vector Machines (SVMs) are supervised learning models used for classification tasks. The NuSVC variant is a variation of SVMs that supports the use of a nu-parameter to control the number of support vectors. In the context of mushroom species classification\cite{svcmush}, NuSVC constructs decision boundaries by identifying the hyperplanes that maximize the margin between different classes in the feature space.

Given a training dataset with features \( \mathbf{x}_i \) and corresponding labels \( y_i \), where \( i = 1, 2, ..., N \), NuSVC aims to find the hyperplane \( \mathbf{w}^T \mathbf{x} + b = 0 \) that separates the instances into different classes while maximizing the margin. This can be formulated as the following optimization problem:

\begin{equation}
	\min_{\mathbf{w}, b, \xi} \frac{1}{2} \| \mathbf{w} \|^2 + C \sum_{i=1}^{N} \xi_i
\end{equation}

subject to:

\begin{equation}
	y_i (\mathbf{w}^T \mathbf{x}_i + b) \geq 1 - \xi_i, \quad \xi_i \geq 0, \quad i = 1, 2, ..., N
\end{equation}

where \( \mathbf{w} \) is the weight vector, \( b \) is the bias term, \( \xi_i \) are slack variables, and \( C \) is the regularization parameter.

\begin{figure}[h]
	\centering
	\includegraphics[height=0.475\linewidth]{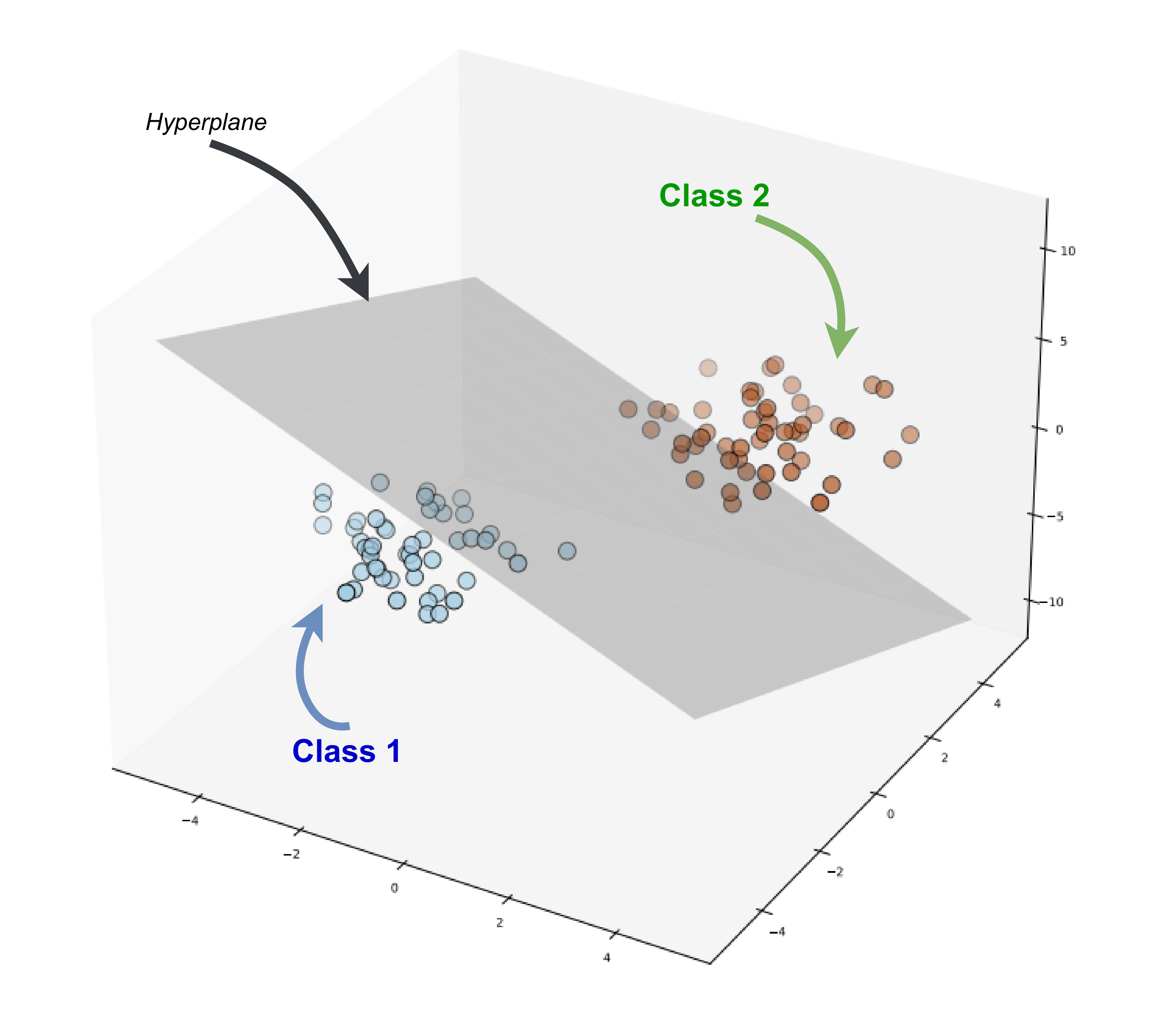}
	\vspace{-0.4cm}
	\caption{Decision Boundary of SVC}
\end{figure}

The decision boundary is determined by the hyperplane \( \mathbf{w}^T \mathbf{x} + b = 0 \), where points on one side of the hyperplane are classified as one class, and points on the other side are classified as the other class.

\subsubsection{XGBClassifier (Extreme Gradient Boosting Classifier)}

XGBoost is an ensemble learning algorithm that utilizes a collection of weak learners, typically decision trees, to construct a strong predictive model. In the context of mushroom species classification\cite{wang2022automatic}, the XGBClassifier constructs decision boundaries by iteratively adding decision trees that minimize a loss function and update the model's predictions.

Ortiz et al. \cite{ortiz2024classification} focused on analyzing fungal characteristics resulting in a toxicity classification models by employing genetic algorithms and LASSO regression. The significant features such as odor, spore print color, and habitat were identified by GALGO, while LASSO regression highlighted odor, gill size, and stalk shape. Variety of machine learning algorithms were used for model training such as logistic regression, k-nearest neighbor (KNN), and ensemble learning namely XGBoost. The result showed that the XGBoost gave the highest AUC value of 0.99 with LASSO-selected features, closely followed by KNN. Selecting the GALGO's features led to a reduced feature set and hence, achieving an AUC of 0.98 for both KNN and XGBoost. For LASSO, Logistic regression gave an accuracy of 0.64, whereas KNN and XGBoost models with GALGO selection showed accuracies over 0.99.

Desingu et al. \cite{desingu2022classification} used deep learning-based approach for feature extraction and gradient boosting ensemble approach for fungi species classification. By effectively utilizing the deep learning architectures namely ResNeXt and EfficientNet through transfer learning on a fungi image dataset, aiming to extract meaningful features for classification. Now by combining the output vectors from deep learning models, XGBoost was employed as a gradient boosting classifier to predict fungi species as accurately as possible. The researchers fine-tuned the XGBoost classifier using a grid-search strategy to optimize hyperparameters, they tunned hyperparameters such as tree depth, learning rates, and the number of decision trees, to enhance the model's performance during training. In their experiment, the set best feature extraction models were selected, giving a maximum macro-averaged F1-Score of 48.96\% on the test data. For validation dataset F1-Score and Accuracy scores were reported as 50.22\% and 85.11\%, respectively. This demonstrated the effectiveness of the approach for accurately predicting the fungal species.

Let \( f_k(\mathbf{x}) \) denote the prediction of the \( k \)-th decision tree, and \( F_k(\mathbf{x}) \) represent the ensemble prediction after \( k \) trees. The XGBoost model aims to minimize the following objective function:

\begin{equation}
	\mathcal{L}(\Phi) = \sum_{i=1}^{N} l(y_i, F_K(\mathbf{x}_i)) + \Omega(\Phi)
\end{equation}

where \( l(y_i, F_K(\mathbf{x}_i)) \) is the loss function measuring the difference between the predicted values and true labels, and \( \Omega(\Phi) \) is the regularization term penalizing the complexity of the model. The decision boundaries in XGBClassifier are constructed based on the ensemble predictions of multiple decision trees. The final decision is made by aggregating the predictions of individual trees, typically using a weighted sum or voting mechanism.

\subsection{Deep Learning Models}

Deep learning models have revolutionized the field of image classification, offering powerful tools for automatic feature extraction and classification. In the context of mushroom species identification, several deep learning architectures have been employed, each with its unique approach to delineating decision boundaries in the feature space. Liu et al. 
\cite{liu2022deep} focused on improving the classification accuracy of shiitake mushrooms by using deep learning technique. The approach they proposed was enhancing the YOLOX deep learning model with channel pruning and knowledge distillation methods. They have used an expanded image dataset and transfer learning technique, YOLOX model was refined to analyze the surface of the shiitake mushroom. It showed an impressive result, with an accuracy (mAP) of 99.96 percentage and frames per second (FPS ) of 57.3856.  After the comparative analysis with other algorithms such as Faster R-CNN, YOLOv3, YOLOv4, and SSD 300, the results showed that the improved YOLOX algorithm outperformed other baseline models in terms of mAP value, and hence showcasing its efficiency and accuracy in classification task.

\subsubsection{Convolutional Neural Networks (CNN)}

Convolutional Neural Networks (CNNs) are a class of deep neural networks particularly well-suited for image classification tasks. CNNs leverage convolutional layers to automatically extract hierarchical features from input images, capturing local patterns and spatial relationships.

Preechasuk et al.\cite{preechasuk2019image} considered 45 types of mushrooms for classification task, the dataset contains both edible and poisonous categories of mushroom, by employing Convolutional Neural Networks (CNNs). The main objective of the research was to mitigate the risk of incidences of illnesses and fatalities after the consumption of toxic mushrooms, often due to the challenging task of differentiating between safe and harmful species. Their CNN model exhibited impressive performance results which are namely, precision of 0.78, recall of 0.73, and F1 score of 0.74. Finally, CCN model accuracy was evaluated, giving an overall classification accuracy of approximately 74\% across the dataset, with a consistent performance across 1,000 training epochs.

\begin{figure}[ht]
	\centering
	\includegraphics[width=0.5\linewidth]{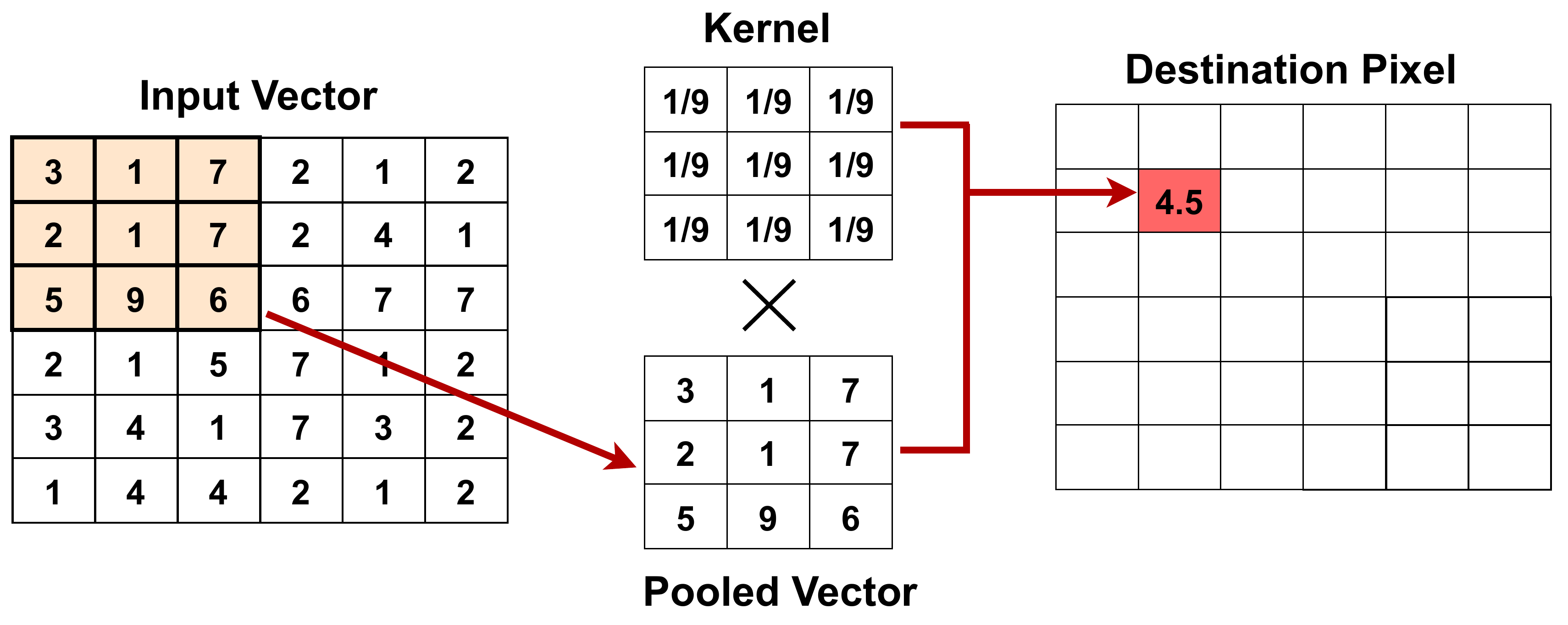}
	\caption{Convolution Operation}
\end{figure}

The fundamental operation in CNNs is the convolution operation\cite{sajedi2019automated}, which involves sliding a filter (also known as a kernel) over the input image and computing the dot product between the filter weights and the corresponding region of the image. Mathematically, the convolution operation can be expressed as:

\begin{equation} Z[i, j] = (f * X)[i, j] = \sum_{m=0}^{F_h-1} \sum_{n=0}^{F_w-1} X[i+m, j+n] \times f[m, n] + b 
	\label{eq:conv}
\end{equation}

Where:
- \( Z \) is the output feature map.
- \( X \) is the input image.
- \( f \) is the filter.
- \( F_h \) and \( F_w \) are the height and width of the filter.
- \( b \) is the bias term. In this equation, \( i \) and \( j \) iterate over the spatial dimensions of the output feature map \( Z \), while \( m \) and \( n \) iterate over the spatial dimensions of the filter \( f \).

An activation function follows the convolution operation, commonly ReLU (Rectified Linear Unit), which introduces non-linearity to the network. The activation function can be expressed as $ f(x) = \max(0, x)$ After applying the activation function, other operations such as pooling may be performed to downsample \cite{devika2021identification} the feature maps and reduce spatial dimensions, thus aiding in translation invariance and computational efficiency. Various types of pooling operations include: 

\begin{enumerate}
	\item \textbf{Max Pooling:} Chooses the maximum value from
	      each subregion of the feature map.\cite{poolmush}
	      \begin{equation} Y[i, j] = \max_{m,n} X[i \times s + m, j \times s + n] \end{equation}
	      	      	      	       
	\item \textbf{Average Pooling:} Chooses the average value from
	      each subregion of the feature map. \cite{poolmush}
	      \begin{equation} Y[i, j] = \frac{1}{k \times l} \sum_{m=0}^{k-1} \sum_{n=0}^{l-1} X[i \times s + m, j \times s + n] \end{equation}
	      	      	      	      
	\item \textbf{Global Average Pooling (GAP):} Chooses the maximum or average
	      value across all the feature maps in a layer. This
	      produces a single scalar value for each feature map. \cite{poolmush}
	      \begin{equation} Y_c = \frac{1}{H \times W} \sum_{i=1}^{H} \sum_{j=1}^{W} X_{ij}^c \end{equation}
	      	      	      	      
	\item \textbf{$\ell^{2}$ Pooling:} Chooses the $\ell^{2}$ norm of each
	      subregion in the feature map. \cite{poolmush}
	      \begin{equation} Y[i, j] = \sqrt{\sum_{m=0}^{k-1} \sum_{n=0}^{l-1} X[i \times s + m, j \times s + n]^2} \end{equation}
\end{enumerate}

In all these expressions:
- \( Y \) is the output of the pooling layer.
- \( X \) is the input to the pooling layer.
- \( s \) is the stride (the amount by which the pooling window shifts).
- \( m \) and \( n \) iterate over the pooling window dimensions.
- \( k \) and \( l \) represent the dimensions of the pooling window for average pooling.
- \( H \) and \( W \) represent the height and width of the input feature map for global average pooling.

\subsubsection{Transfer Learning}

Consider the analogy of acquiring skills, such as riding a bicycle, which entails developing fundamental abilities like balance, coordination, and spatial awareness. Subsequently, when attempting to acquire a related skill, such as riding a skateboard, it becomes evident that these foundational abilities can be transferred, facilitating the learning process.

Transfer learning could be defined as a technique in deep learning where a model trained on one task is leveraged for a different but related task. In the context of mushroom species identification, pre-trained CNN models\cite{tan2019efficientnet} such as ResNet, VGG16, Inception-V3, and EfficientNet-B5, which have been trained on large-scale image datasets like ImageNet, are fine-tuned on mushroom image data.

Mathematically, transfer learning involves initializing the weights of the pre-trained model with learned parameters from the original task and then fine-tuning these weights on the new task. The fine-tuning process typically involves adjusting the model's parameters through backpropagation using a smaller learning rate to adapt the model to the specifics of the new dataset.

By transferring knowledge from pre-trained models, transfer learning accelerates the training process and enhances the generalization capabilities of the model for mushroom species classification.

Zahan et al. \cite{zahan2021deep} task was to classifying mushrooms into edible, inedible, and poisonous buckets by employing deep learning models like InceptionV3, VGG16, and Resnet50. For the initial run, raw and unprocessed images were used for classification, and as expected the results were unsatisfactory.  Further by changing the dataset to highly contrast-enhanced images, it significantly improved the classification accuracy. Among the different CNN architectures assessed, VGG16 achieved the highest test accuracy of 84.2 percentage with raw images, followed by InceptionV3 at 82.31 percentage and Resnet50 at 53.25 percentage. In their result InceptionV3 model outperformed the others models, achieving the highest accuracy of 88.40 percentage. The paper showcased the effectiveness of transfer learning with pre-trained CNN models for quicker and more efficient classification tasks.

\subsubsection{Vision Transformer}

The Vision Transformer (ViT) is a novel architecture that applies the Transformer model, originally developed for natural language processing tasks, to the domain of computer vision. Unlike traditional convolutional neural networks (CNNs), which rely on convolutional layers for feature extraction, ViT leverages self-attention mechanisms to capture global dependencies within images \cite{wang2022automatic}.
At its core, the ViT architecture consists of multiple layers of self-attention modules, also known as Transformer blocks. Each Transformer block comprises two main components: the multi-head self-attention mechanism and the position-wise feedforward network. 

The transformer network starts by decomposing an image into patches and pre-processing the
set of patches to map each one into a vector. The initial set of patches is enhanced with an extra vector of the same size as the patches, called class embedding. This class
embedding vector is used at the end of the network, to feed into a fully connected layer that yields the output. One trainable vector called positional embedding is also included, which is added to each vector \cite{chen2021chasing}. 

In addition to the self-attention mechanism and position-wise feedforward network, ViT incorporates learnable positional embeddings to provide spatial information to the model. These embeddings encode the position of each input patch within the image, enabling the model to distinguish between different regions.

The multi-head self-attention mechanism allows the model to attend to different parts of the input image simultaneously, enabling it to capture long-range dependencies and relationships between image patches. This mechanism computes attention scores between pairs of input patches, which are then weighted and aggregated to generate contextualized representations for each patch.

Mathematically, the self-attention mechanism can be expressed as follows:

\begin{equation}
	\textit{{Attention}}(Q, K, V) = \textit{{softmax}}\left(\frac{QK^T}{\sqrt{d_k}}\right)\cdot V
\end{equation}

where \(Q\), \(K\), and \(V\) denote the query, key, and value matrices, respectively, and \(d_k\) represents the dimensionality of the key vectors. The softmax operation normalizes the attention scores, ensuring that they sum up to one.

Following the self-attention mechanism, the position-wise feedforward network applies a fully connected layer separately to each position in the input sequence, enabling the model to capture non-linear relationships between features. This network consists of two linear transformations with a GeLU activation function in between:

\begin{equation}
	\text{{FFN}}(x) = \text{{GeLU}}(xW_1 + b_1)W_2 + b_2
\end{equation}

where \(W_1\), \(W_2\), \(b_1\), and \(b_2\) are learnable parameters. The GELU (Gaussian Error Linear Unit) activation function is defined as:

\begin{equation}
	\text{GELU}(x) = x \cdot \Phi(x)
\end{equation}

where \( \Phi(x) \) is the standard Gaussian cumulative distribution function (CDF), given by:

\begin{equation}
	\Phi(x) = \frac{1}{2} \left(1 + \text{erf}\left(\frac{x}{\sqrt{2}}\right)\right)
\end{equation}

Here, \( \text{erf}(x) \) represents the error function. The GELU activation function introduces non-linearity to the network and exhibits state-of-the-art performance in Transformer-based architectures like the Vision Transformer (ViT). This activation function addresses issues such as vanishing gradients and dead neurons commonly encountered when using ReLU in a network. Studies have shown that Vision Transformers (ViTs) can capture global relationships in images, which makes them a more scalable and flexible alternative to Convolutional Neural Networks (CNNs). However, the training of ViTs requires significant computational resources, which can pose limitations in certain situations.

\subsection{Hybrid Quantum Deep Learning}

It involves integrating a small-scale quantum circuit with a classical neural network, as illustrated in the standard two-step approach akin to classical machine learning\cite{engineer}:

\begin{enumerate}
	\item \textbf{Selection of PQC architecture (Ansatz):} The designer chooses the architecture of a Parametric Quantum Circuit (PQC) by specifying a sequence of parametrized quantum gates, similar to selecting a neural network architecture. The PQC's operation is defined by a unitary matrix \(U(\theta)\), dependent on a vector of free parameters \(\theta\). This selection ideally aligns with known quantum algorithmic architectures suitable for the given problem\cite{engineer}.
	\item \textbf{Parametric Optimization with backpropagation:} The PQC implementing the unitary \(U(\theta)\) is connected to a classical neural network with a classical optimizer. The optimizer updates the parameter vector \(\theta\) based on measurements of the quantum state produced by the PQC, aiming to minimize a cost function, often related to training data, similar to classical machine learning's gradient descent-based optimization\cite{wang2022quantumnas}.
\end{enumerate}

The integrated approach of quantum machine learning offers significant advantages over solely handcrafting quantum algorithms. It addresses key challenges like quantum resource limitations and facilitates the automatic design of efficient quantum algorithms. A crucial step in this process is data encoding, where the input \(x\) is encoded into the quantum state produced by the Parametric Quantum Circuit (PQC).

\subsubsection{Angle Encoding}

Angle encoding efficiently converts input \(x\) into a quantum state using a unitary transformation \(U(x, \theta)\), creating a quantum embedding of the classical input. The PQC \(U(x, \theta)\) alternates between unitaries dependent on \(x\) and those dependent on the model parameter vector \(\theta\), potentially entering data \(x\) multiple times or progressively at each layer. The unitaries, \(V_l(x)\) and \(W_l(\theta)\), consist of single-qubit rotations and possibly parametrized entangling gates, with \(V_l(x)\) typically selected for expressivity and \(W_l(\theta)\) for training facilitation. Angle encoding requires at least \(M\) qubits to encode an input \(x\) containing \(M\) real numbers.\cite{engineer}

\begin{equation}
	U(\theta) = W_L(\theta)V_L(x) \cdot W_{L-1}(\theta)V_{L-1}(x) \cdot \ldots \cdot W_1(\theta)V_1(x)
\end{equation}

The output state of the PQC $U(x, \theta)$, i.e.,

\begin{equation}
	|\psi(x, \theta)\rangle = U(x, \theta)|0\rangle
\end{equation}

\subsubsection{Beyond Angle Encoding}

\begin{enumerate}
	\item \textbf{Amplitude Encoding:} This analog strategy maps each entry \(x_k\) of \(x\) into the \(k\)-th amplitude of a quantum state, providing exponential efficiency in qubit usage compared to angle encoding. However, its computational complexity remains linear in the size \(M\) of the input.\cite{engineer}
	      	      	          
	      The quantum embedding for data vector \(x\) is:
	      \begin{equation}
	      	|\psi(x)\rangle = \sum_{k=0}^{M-1} f(x_k)|k\rangle
	      \end{equation}
	      	      	          
	\item \textbf{Basis Encoding:} Unlike angle and amplitude encoding, this digital strategy encodes an entire data set \(\mathcal{D}\) of data points \(x\) into a single quantum state by converting each \(x\) into a binary string \(x_b\) and mapping \(\mathcal{D}\) to a superposition of these strings.\cite{engineer}
	      \begin{equation}
	      	|\psi(\mathcal{D})\rangle = \frac{1}{Z} \sum_{x \in \mathcal{D}} |x_b\rangle,
	      \end{equation}
\end{enumerate}

\subsection{Quanvolutional Neural Networks}

The concept of quanvolutional neural networks (QNNs), introduced by Henderson et al. [2020] \cite{henderson2020quanvolutional}, 
draws inspiration from classical 2D convolutions. Similar to classical convolutional layers, a quanvolutional layer generates feature maps from input tensors through localized transformations. However, instead of conducting element-wise matrix-matrix multiplications, the quanvolutional layer initially encodes an image patch into a quantum state |Ii and processes that state through a sequence of two-qubit gates (e.g., CNOT) and parameterized one-qubit gates (e.g., rotations Ra($\theta$) around the X, Y, and Z axes of the Bloch sphere). Figure 4 illustrates the core concept of the QNN approach.

The input data is encoded into a quantum state by mapping each element of the input vector to the rotation angle of a qubit around a Pauli basis (X, Y, or Z). Mathematically, the encoding algorithm can be expressed as follows:

\begin{equation}
	\mathbf{\psi}(x_i) = e^{-i\dfrac{x^0_i}{2}\theta_{j_1}} \otimes \cdots \otimes e^{-i\dfrac{x^{n-1}_i}{2}\theta_{j_n}}
	\label{mlp}
\end{equation}

Here, $\theta_{j_k} \in \{ \theta_x, \theta_y, \theta_z \}$, $\mathbf{x}_i = (x^0_{i}, x^1_{i}, \ldots, x^{n-1}_{i})$ is n-dimensional input data, and $\otimes$ combines individual qubit state spaces. Compared to the entanglement encoding scheme proposed in \cite{wang2022quantumnas},
this scheme assigns each input information to a separate qubit, minimizing the impact of errors on encoding circuits and maximizing fidelity transformation between classical input data and quantum input feature maps and the linear form of the above expression does not necessarily mean the underlying gate or function is linear.

\begin{figure}[h]
	\centering
	\includegraphics[width=1\linewidth]{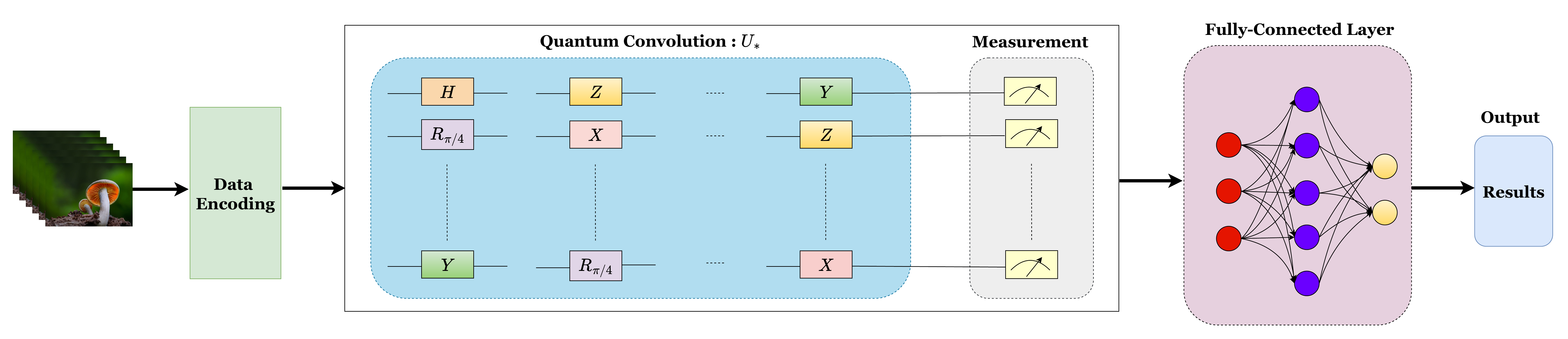}
	\vspace{-0.7cm}
	\caption{Quanvolutional Neural Network}
\end{figure}

As a hybrid quantum-classical algorithm tailored for the Noisy Intermediate-Scale Quantum (NISQ) era, QNNs accommodate the limited width of quantum hardware. While operating on entire images is impractical due to hardware constraints, processing smaller patches, as in a convolution, becomes feasible with the current number of available qubits. Furthermore, the encoding of small filter sizes circumvents the need for Quantum Random Access Memory (QRAM) technology, which is still under development. QNNs retain the advantages of convolutions by learning local patterns in a translationally invariant manner across the entire image.

\begin{figure}[h]
	\centering
	\includegraphics[height=0.4\linewidth]{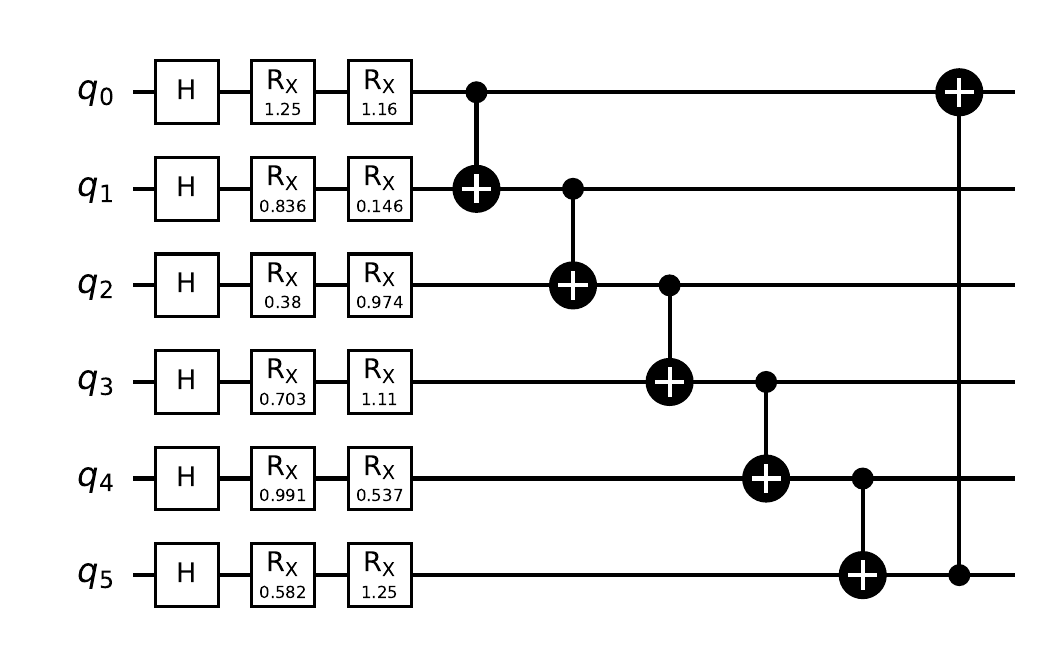}
	\vspace{-0.5cm}
	\caption{The Quantum circuit utilized in the QMViT architecture constructed with H, RX, and CNOT gates}
\end{figure}
        
\subsection{Hybrid Quantum-Classical Vision Transformer}

Our research methodology is inspired by Unlu, E.B \textit{et al.} \cite{unlu2024hybrid}, $2024$ in hybrid quantum vision transformer development. The Hybrid Quantum-Classical Vision Transformer (HQViT) combines classical and quantum components to perform image classification tasks
\cite{xue2024end}
. Inspired by the classical Vision Transformer (ViT) architecture, HQViT incorporates quantum processing elements to enhance its capabilities. The key component of HQViT is the Hybrid Self-Attention mechanism, which integrates classical and quantum operations.

The Hybrid Quantum Self-Attention mechanism in HQViT \cite{pasqualimeasurements, cherrat2022quantum}
combines classical multi-head self-attention with quantum processing to capture both local and global dependencies in the input image. Then, this output is fed into a quantum multilayer perceptron of two Linear layers and one Quantum Layer, and the result is passed through a Layer Normalization layer followed by a final Linear Layer of 100 out features. This mechanism follows a step-by-step procedure outlined below:

\begin{algorithm}[h]
	\label{alg:qvitmush}
	\caption{Quantum Vision Transformer}
	\begin{algorithmic}
		\setstretch{1} 
		\STATE \textbf{Inputs:} Image data (training, validation)
		\STATE Initialise QMViT model
		\FOR{each epoch in training}
		\FOR{each batch in training data}
		\STATE Load and preprocess batch of images
		\STATE Apply data augmentation (e.g., random rotation, sharpness adjustment)
		\STATE Create Patch Embeddings of the batch via Convolution Operations \eqref{eq:conv}
		\STATE Initialise Quantum Transformer Blocks for QMViT.
		\STATE Apply angle encoding to the features \eqref{eq:data-encoding}
		\STATE Express the parameters as PQCs for the Query\eqref{Q}, Key\eqref{K}, Value\eqref{V}, and Output\eqref{O} tensors.
		\STATE Generate tunable parameters for the Quantum Multi-head Self-Attention module.
		\STATE Update Quantum Transformer Blocks using input images and trainable parameters. \eqref{A}
		\STATE Perform Layer Normalization for global features.
		\STATE Apply Quantum Multilayer Perceptron\eqref{mlp} as the Feed Forward Network Module.
		\STATE Apply a linear classifier head to make final predictions.
		\STATE Compute loss and update parameters using Adam optimizer
		\ENDFOR
		\STATE Assess QMViT model on validation data
		\ENDFOR
		\STATE Compute performance metrics on test data
	\end{algorithmic}
\end{algorithm}

\begin{figure}[htbp]
	\begin{minipage}{0.8\textwidth}
		\centering
		\includegraphics[width=.7\linewidth]{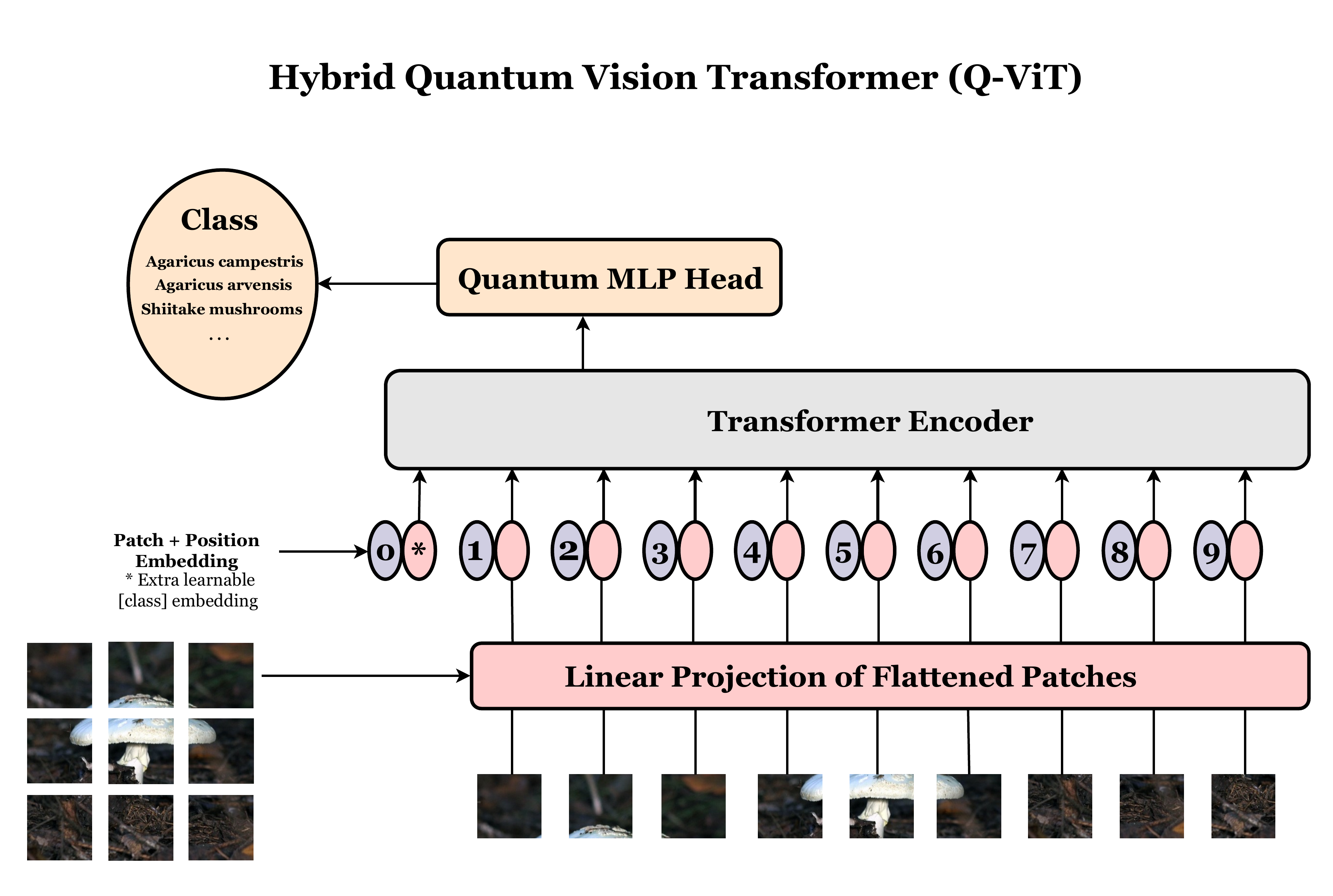}
	\end{minipage}
	\hspace{-2.5cm}
	\begin{minipage}{0.3\textwidth}
		\centering
		\includegraphics[height=1.2\linewidth]{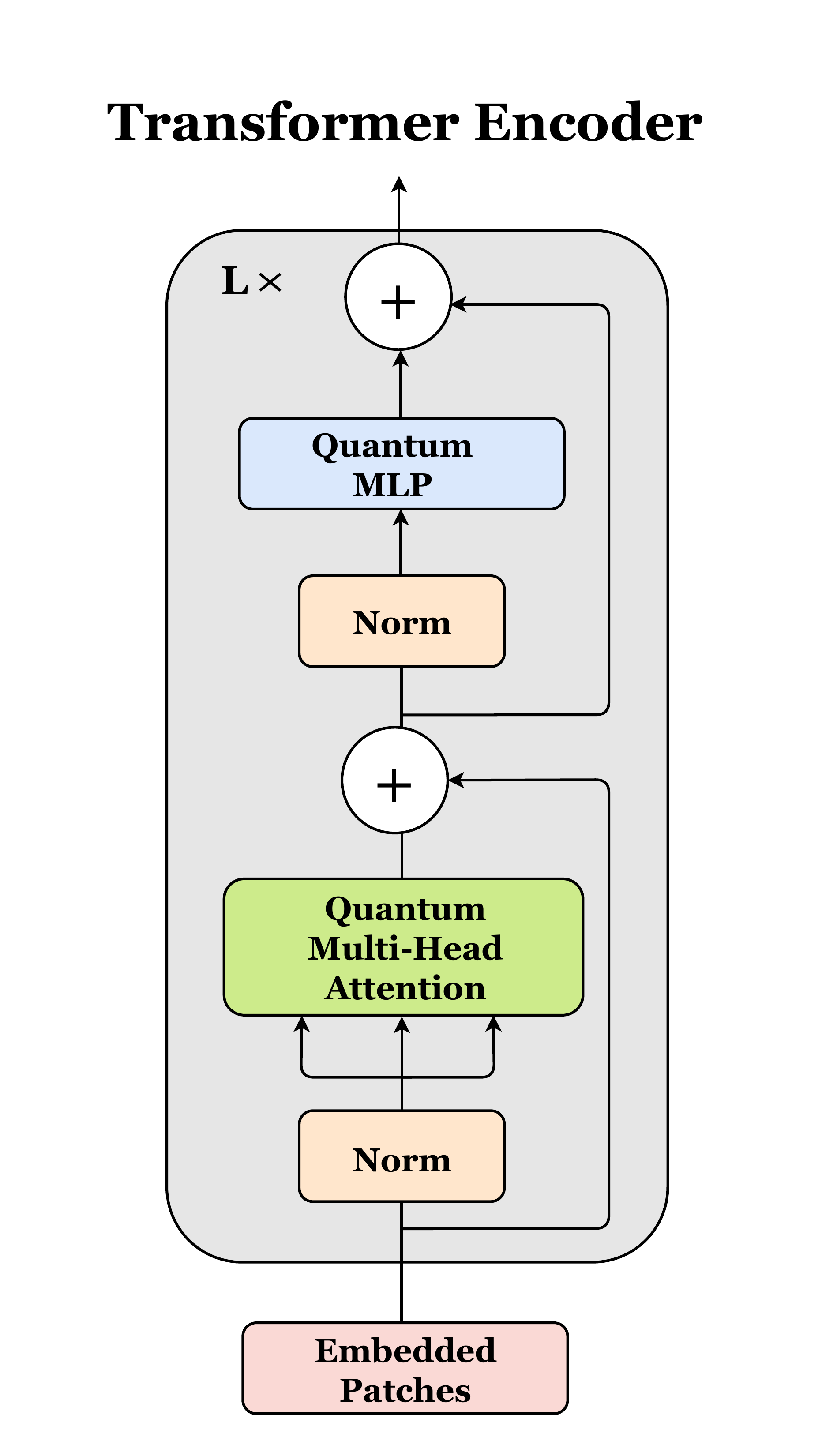}
	\end{minipage}
	\caption{Hybrid Classical-Quantum Vision Transformer (Q-ViT)}
\end{figure}

\begin{enumerate}
	\item \textbf{Data Encoding}: Each input row $x_i$ of each Patch embedding $X$ is encoded into a quantum state using a data loader operator $U^\dagger(x_i)$, defined as:
	      \begin{equation}
	      	|x_i\rangle \equiv U^\dagger(x_i)|0\rangle = \bigotimes_{j=1}^{dh} R^\dagger_x(x_{ij})H|0\rangle
	      	\label{eq:data-encoding}
	      \end{equation}
	      where $H$ represents the Hadamard gate and $R_x$ is a parameterized rotation around the x-axis.	          
	\item \textbf{Key and Query Operations}: For each input row $x_i$, the key operator $K^\dagger(\theta_K)$ and query operator $Q^\dagger(\theta_Q)$ are applied to obtain the key vector $K$ and query vector $Q$, respectively:
	      \begin{align}
	      	K_i & = \langle x_i | K^\dagger(\theta_K)Z_0 K(\theta_K) | x_i \rangle \label{K} \\
	      	Q_i & = \langle x_i | Q^\dagger(\theta_Q)Z_0 Q(\theta_Q) | x_i \rangle           
	      	\label{Q}
	      \end{align}
	      where $Z_0$ represents a spin measurement of the qubit on the z-direction.
	      	      	          
	\item \textbf{Attention Matrix Calculation}: The attention matrix $A$ is computed using the key and query vectors:
	      \begin{equation}
	      	A_{ij} = -(Q_i - K_j)^2
	      	\label{A}
	      \end{equation}
	      	      	          
	\item \textbf{Value Operation}: Each row of the image is passed through a value operator $V^\dagger(\theta_V)$ to obtain the value matrix $V$:
	      \begin{equation}
	      	V_{ij} = \langle x_i | V^\dagger(\theta_V)Z_j V(\theta_V) | x_i \rangle
	      	\label{V}
	      \end{equation}
	      	      	          
	\item \textbf{Hybrid Attention Head}: Finally, the Hybrid Attention Head applies the softmax function to the attention matrix $A$ divided by $\sqrt{dh}$, followed by multiplication with the value matrix $V$:
	      \begin{equation}
	      	\text{Hybrid Attention Head: } \text{SoftMax}\left(\frac{A}{\sqrt{dh}}\right) \cdot V
	      	\label{O}
	      \end{equation}
\end{enumerate}

The detailed process illustrated in Figure 6 elucidates the architecture and operational intricacies of the Hybrid Quantum Vision Transformer, emphasizing the individual components and their interactions in Mushroom Classification. Moreover, Algorithm 1 provides a sequential account of the implementation.

\section{Results}

We evaluated the performance of our classical Vision Transformer (ViT) Model with Transfer Learning, Quanvolutional Neural Network (QNN) with Transfer Learning, and hybrid-quantum ViT with no inherent weights model on the same dataset described in Section 3.1. The results demonstrated that the \texttt{ViT} model achieved promising results on the mushroom classification task. The model was trained for $10$ epochs using the Adam optimizer with a batch size of $32$ utilizing Keras, as detailed in Table~\ref{tab:hyperparameters}. These hyperparameters were selected based on preliminary experiments and empirical observations to strike a balance between model complexity and generalization performance.

\begin{figure}
    \centering
    \includegraphics[height=0.5\linewidth]{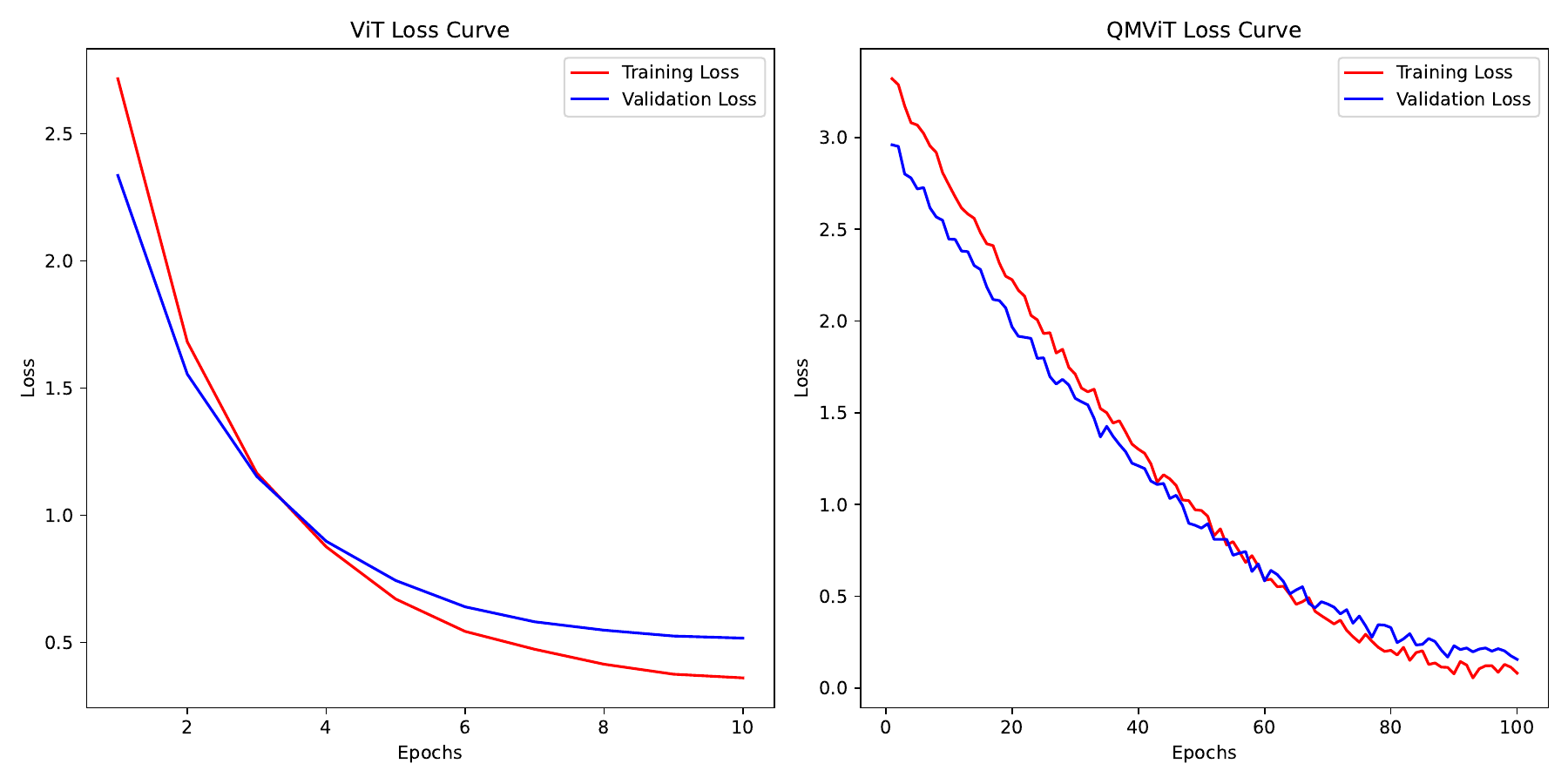}
    \caption{Training \& Validation loss curves of best ViT \& QMViT for 4 Qubits and 1 Layer}
    \label{fig:loss-curve}
\end{figure}

\begin{table}[htpb]
	\centering
	\caption{Hyperparameters and training details for the \texttt{ViT} model.}
	\vspace{0.2cm}
	\begin{tabularx}{0.4\textwidth}{@{}lX@{}}
		\toprule
		\textbf{Hyperparameter} & \textbf{Value}                       \\
		\midrule
		Model Architecture      & \texttt{Google's 224 ViT Base Patch} \\
		Number of Epochs        & $10$                                 \\        
		Batch Size              & $32$                                 \\
		\bottomrule
	\end{tabularx}
	\label{tab:hyperparameters}
\end{table}

The loss curves in Fig.~\ref{fig:loss-curve} show that the training loss decreases steadily over time while the validation loss decreases more slowly, which suggests that the model is well-fitted on both the training and validation data.

We performed experiments on the hybrid quantum-classical vision transformer using pennylane \cite{pennylane} under different scenarios, namely qubit count and hidden layers. The results in Tab.~\ref{tab:Q_res} show that a dense encoding scheme of image features using fewer qubits performed better than using more qubits and sparser representations, and this is evident as the model with $4$ qubits, with an accuracy of $92.33\%$ performed best compared to that with $8$ qubits, with an accuracy of $87.66\%$.

Careful analysis of quantum techniques suggests that better representation, deeper search space, state entanglement, and qubit parallelism are key factors contributing to the supremacy of quantum circuits for classification tasks. In our experiments, we observed the same with a smaller system with 4 qubits which makes our model QMViT outperform the classical ViT. These results suggest that with more advanced features and complex quantum systems, we can achieve a real quantum advantage in solving problems such as the edibility detection of mushrooms in highly imbalanced datasets.

\begin{table*}[htpb]
	\centering
	\caption{Performance test metrics of QMViT. The table presents results for various configurations, including the number of qubits and layers.}
	\vspace{0.2cm}
	\begin{tabular}{|c|c|c|c|c|c|}
		\hline
		\textbf{Num Qubits} & \textbf{Layers} & \textbf{Accuracy (\%)} & \textbf{F1 score} \\
		\hline
		\hline 
		4                   & 1               & 92.33                  & 0.87              \\
		8                   & 1               & 87.56                  & 0.81              \\
		4                   & 2               & 89.66                  & 0.83              \\
		8                   & 2               & 85.33                  & 0.77              \\
		\hline
	\end{tabular}
	\label{tab:Q_res}
\end{table*}

The below gives the entire benchmark of all the different standards of models based on different datasets as well as a prospective overview of how our model fares with the existing model.

\begin{figure}[htbp]
    \centering
    \includegraphics[height=0.6\linewidth] {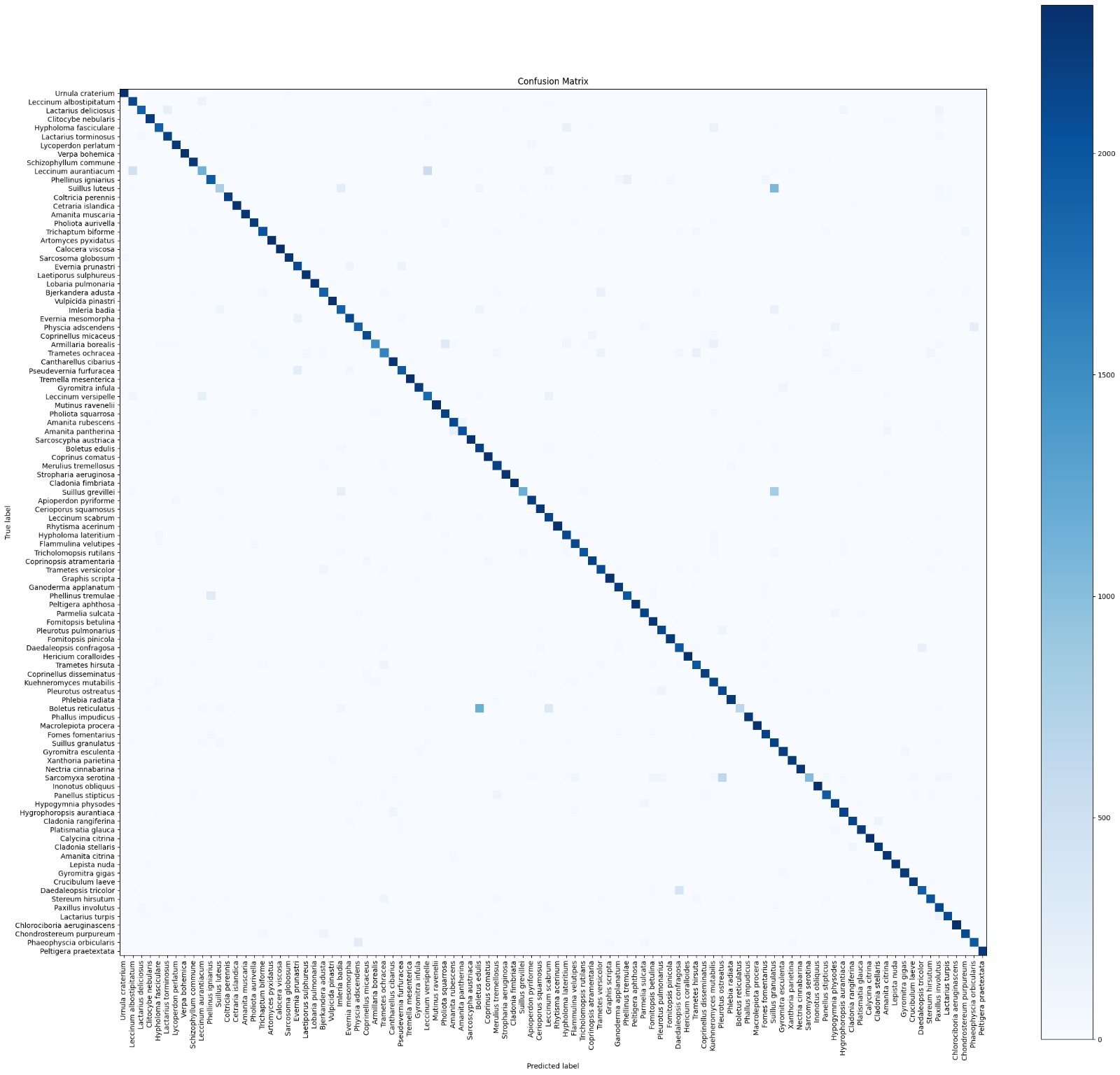}
    \caption{Confusion Matrix of QMViT}
    \label{fig:confmat}
\end{figure}

\begin{table}[htbp]
	\centering
	\caption{Model Performance Summary in terms of Mushroom Classification}
	\label{tab:model_performance}
	\begin{tabular}{@{}lcccc@{}}
		\toprule
		Model Name                                  & Number of Classes & Number of Images & Performance           \\ \midrule
		SVC+FT-IR Scan \cite{gao2020classification} & 2                 & 363              & 99.06\% Acc.          \\
		YOLOv5 \cite{cengil2021poisonous}           & 8                 & 644              & 77.8\% mAP            \\
		KNN \cite{cengil2021poisonous}              & 2                 & 380              & 87\% - 94\% Acc.      \\
		CNN \cite{preechasuk2019image}              & 45                & 8556             & 74\% Acc.             \\
		VGG16 \cite{zahan2021deep}                  & 45                & 8190             & 84.2\% Acc.           \\
		InceptionV3 \cite{zahan2021deep}            & 45                & 8190             & 88.40\% Acc.          \\
		ResNet \cite{zahan2021deep}                 & 45                & 8190             & 58\% Acc.             \\ 
		ViT                                         & 100               & 77400            & 88.85\% Acc.          \\
		QNN(Transfer Learning)                      & 100               & 77400            & 86.33\% Acc.          \\
		\textbf{QMViT (Proposed)}                   & \textbf{100}      & \textbf{77400}   & \textbf{92.33\%} Acc. \\ \bottomrule
	\end{tabular}
\end{table}

An in-depth analysis of the confusion matrix \ref{fig:confmat} generated by the proposed QMViT architecture has a compelling pattern. Despite instances where the specific mushroom species was misclassified if the ground truth label denoted an edible variety, the misclassification consistently fell within the edible category. Conversely, when the ground truth label corresponded to an inedible mushroom species, the misclassification invariably belonged to the inedible category as well. This phenomenon can be attributed to the quantum layers' capability to discern and exploit an inherent separable hyper-geometric structure embedded within the data manifold. The dimensionality of the quantum Hilbert space endowed the system with a heightened capacity to effectively handle high-dimensional data representations. This capability, which remains a formidable challenge for classical approaches, allowed the QMViT architecture to achieve an impressive 99.24\% accuracy compared to classical ViT achieving an accuracy of 97.62\% in classifying mushroom edibility. 

\section{Conclusion}

Our research has demonstrated that a hybrid quantum-classical vision transformer has the potential to detect different species of mushrooms in a large-scale dataset. This is a significant achievement as quantum deep learning techniques in the NISQ era are still in progress regarding massive datasets. We compared the performance of QMViT with ViT and found that QMViT outperformed ViT in terms of accuracy and efficiency. Our experiments on a real-world financial dataset show that QMViT can detect edibility with remarkable accuracy of $99.24\%$, which is significantly higher than ViT's accuracy of $97.62\%$.

Our research shows that QMViT has the potential to revolutionize mushroom detection and ensure food safety. It also highlights the need for further research to address the challenges of implementing quantum computing in real-world applications.

\bibliographystyle{unsrt}
\bibliography{references}

\end{document}